\documentclass[11pt,a4paper]{article}
\usepackage[hyperref]{acl2021}
\usepackage{times}
\usepackage{latexsym}

\usepackage{subcaption}
\usepackage{microtype}
\usepackage{graphicx}
\usepackage{multirow}
\usepackage{pifont}

\usepackage{microtype}

\aclfinalcopy % Uncomment this line for the final submission
\title{StructuralLM: Structural Pre-training for Form Understanding}

\author{Chenliang Li, Bin Bi, Ming Yan, Wei Wang,\\
\textbf{Songfang Huang, Fei Huang, Luo Si} \\
  Alibaba Group \\
  {\tt \{lcl193798, b.bi, ym119608, hebian.ww\}@alibaba-inc.com} \\
  {\tt \{songfang.hsf, f.huang, luo.si\}@alibaba-inc.com}}

\date{}

\begin{document}
\maketitle
\begin{abstract}
Large pre-trained language models achieve state-of-the-art results when fine-tuned on downstream NLP tasks. However, they almost exclusively focus on text-only representation, while neglecting cell-level layout information that is important for form image understanding. In this paper, we propose a new pre-training approach, StructuralLM, to jointly leverage cell and layout information from scanned documents. Specifically, we pre-train StructuralLM with two new designs to make the most of the interactions of cell and layout information: 1) each cell as a semantic unit; 2) classification of cell positions. The pre-trained StructuralLM achieves new state-of-the-art results in different types of downstream tasks, including form understanding (from 78.95 to 85.14),  document visual question answering (from 72.59 to 83.94) and document image classification (from 94.43 to 96.08).
\end{abstract}

\section{Introduction}
\begin{figure*}[t]
     \centering
     \begin{subfigure}{0.23\textwidth}
         \centering
         \includegraphics[width=\textwidth]{6.png}
         \caption{}
         \label{fig:form}
     \end{subfigure}
     \hfill
     \begin{subfigure}{0.23\textwidth}
         \centering
         \includegraphics[width=\textwidth]{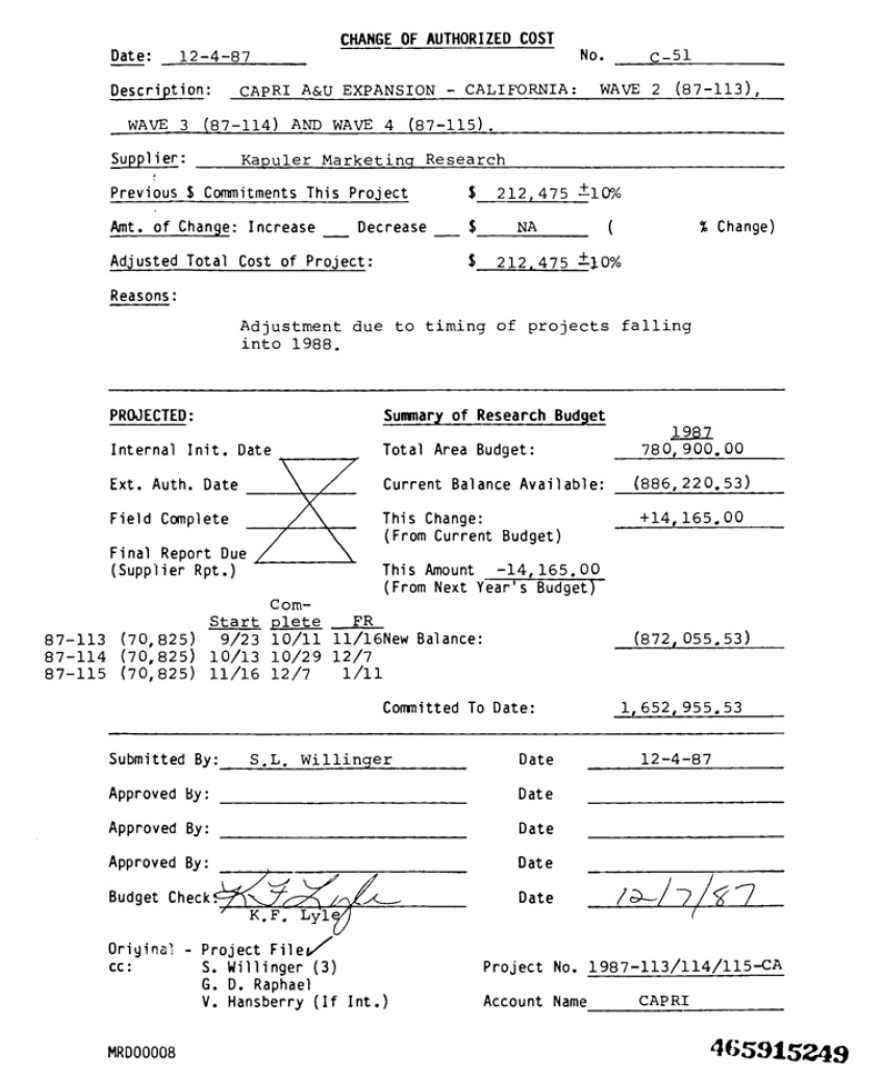}
         \caption{}
         \label{fig:table}
     \end{subfigure}
     \hfill
     \begin{subfigure}{0.23\textwidth}
         \centering
         \includegraphics[width=\textwidth]{3.png}
         \caption{}
         \label{fig:form}
     \end{subfigure}
     \hfill
     \begin{subfigure}{0.23\textwidth}
         \centering
         \includegraphics[width=\textwidth]{4.png}
         \caption{}
         \label{fig:table}
     \end{subfigure}     
     \caption{Scanned images of forms and tables with different layouts and formats.}
     \label{fig:example}
\end{figure*}

Document understanding is an essential problem in NLP, which aims to read and analyze textual documents. In addition to plain text, many real-world applications require to understand scanned documents with rich text. As shown in Figure~\ref{fig:example}, such scanned documents contain various structured information, like tables, digital forms, receipts, and invoices. The information of a document image is usually presented in natural language, but the format can be organized in many ways from multi-column layout to various tables/forms.

Inspired by the recent development of pre-trained language models \cite{devlin-etal-2019-bert,liu2019roberta,wang2019structbert} in various NLP tasks, recent studies on document image pre-training \cite{zhang2020trie,layoutlm} have pushed the limits of a variety of document image understanding tasks, which learn the interaction between text and layout information across scanned document images. 

\citet{layoutlm} propose LayoutLM, which is a pre-training method of text and layout for document image understanding tasks. It uses 2D-position embeddings to model the word-level layout information. However, it is not enough to model the word-level layout information, and the model should consider the cell as a semantic unit. It is important to know which words are from the same cell and to model the cell-level layout information. For example, as shown in Figure~\ref{fig:example} (a), which is from form understanding task\cite{FUNSD}, determining that the "LORILLARD"  and the "ENTITIES" are from the same cell is critical for semantic entity labeling. The "LORILLARD ENTITIES" should be predicted as \emph{Answer} entity, but LayoutLM predicts "LORILLARD" and "ENTITIES" as two separate entities.

The input to traditional natural language tasks is usually presented as plain text, and text-only models need to obtain the semantic representation of the input sentences and the semantic relationship between sentences. In contrast, document images like forms and tables are composed of cells that are recognized as bounding boxes by OCR. As shown in Figure~\ref{fig:example}, the words from the same cell generally express a meaning together and should be modeled as a semantic unit. This requires a text-layout model to capture not only the semantic representation of individual cells but also the spatial relationship between cells. %Therefore, the proposed model in this paper follows the direction, and we explore how to further improve the interactions of cells and layout information.

In this paper, we propose StructuralLM to jointly exploit cell and layout information from scanned documents. Different from previous text-based pre-trained models \cite{devlin-etal-2019-bert,wang2019structbert} and LayoutLM\cite{layoutlm}, StructuralLM uses cell-level 2D-position embeddings with tokens in a cell sharing the same 2D-position. This makes StructuralLM aware of which words are from the same cell, and thus enables the model to derive representation for the cells. In addition, we keep classic 1D-position embeddings to preserve the positional relationship of the tokens within every cell. We propose a new pre-training objective called cell position classification, in addition to the masked visual-language model. Specifically, we first divide an image into N areas of the same size, and then mask the 2D-positions of some cells. StructuralLM is asked to predict which area the masked cells are located in. In this way, StructuralLM is capable of learning the interactions between cells and layout. We conduct experiments on three benchmark datasets publicly available, all of which contain table or form images.
%The first is the FUNSD dataset\cite{FUNSD}, a form understanding task, which contains 199 fully annotated forms. The second is the DocVQA dataset\cite{DocVQA}, a visual question answering task, which contains 50,000 questions defined on over 12,000 images, including tables, forms, etc. The third is the RVL-CDIP dataset\cite{rvlcdip}, a document image classification task, consists of 400,000 images in 16 classes.
Empirical results show that our StructuralLM outperforms strong baselines and achieves new state-of-the-art results in the downstream tasks. In addition, StructuralLM does not rely on image features, and thus is readily applicable to real-world document understanding tasks.

We summarize the major contributions in this paper as follows:
\begin{itemize}
    \item We propose a structural pre-trained model for table and form understanding. It jointly leverages cells and layout information in two ways: cell-level positional embeddings and a new pre-training objective called cell position classification.
    \item StructuralLM significantly outperforms all state-of-the-art models in several downstream tasks including form understanding (from 78.95 to 85.14),  document visual question answering (from 72.59 to 83.94) and document image classification (from 94.43 to 96.08).
\end{itemize}

\section{StructuralLM}
\begin{figure*}[t]
     \centering
     \includegraphics[width=1.0\textwidth]{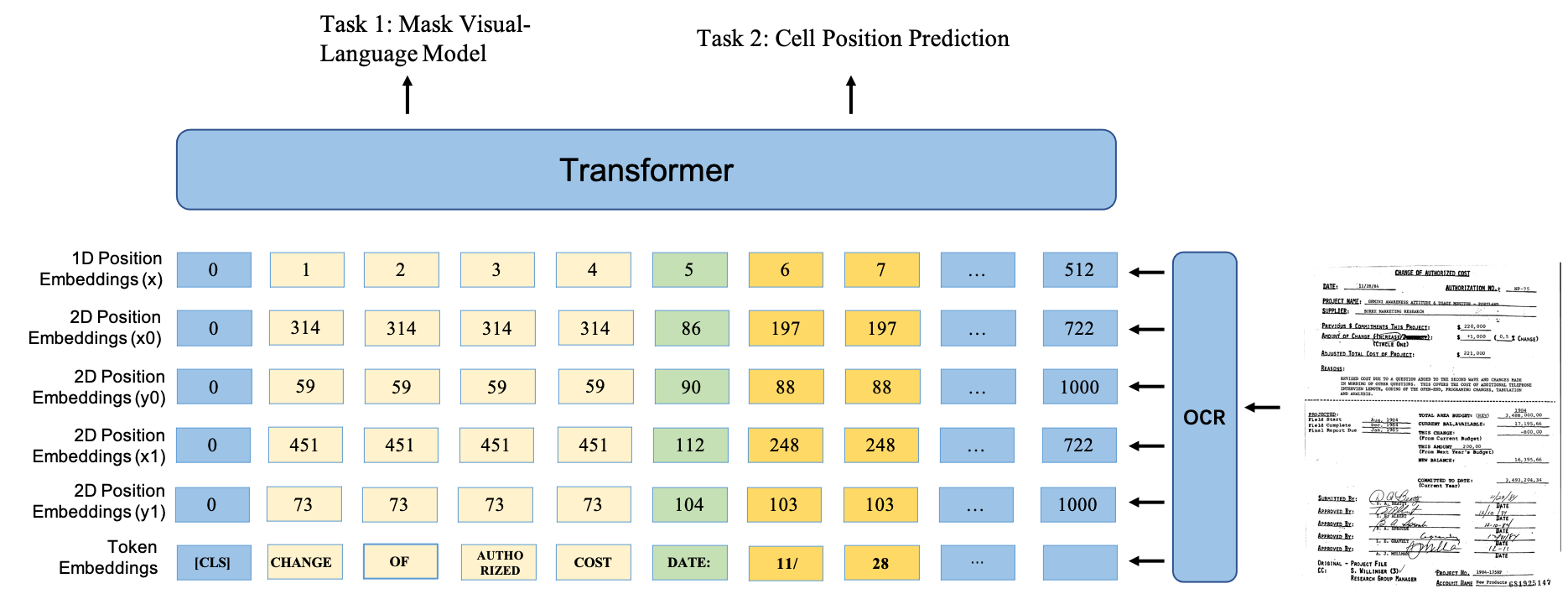}
     \caption{The overall framework of StructuralLM. The input words with the same color background are from the same cell, and the corresponding 2D-positions are also the same.}
     \label{fig:framework}
\end{figure*}

We present StructuralLM, a self-supervised pre-training method designed to better model the interactions of cells and layout information in scanned document images. The overall framework of StructuralLM is shown in Figure~\ref{fig:framework}. Our approach is inspired by LayoutLM\cite{layoutlm}, but different from it in three ways. First, we use cell-level 2D-position embeddings to model the layout information of cells rather than word-level 2D-position embeddings. We also introduce a novel training objective, the cell position classification, which tries to predict the position of the cells only depends on the position of surrounding cells and the semantic relationship between them. Finally, StructuralLM retains the 1D-position embeddings to model the positional relationship between tokens from the same cell, and removes the image embeddings in layoutlm that is only used in the downstream tasks.
\subsection{Model Architecture}
The architecture overview of StructuralLM is shown in Figure~\ref{fig:framework}. To take advantage of existing pre-trained models and adapt to document image understanding tasks, we use the BERT\cite{devlin-etal-2019-bert} architecture as the backbone. The BERT model is an attention-based bidirectional language modeling approach. It has been verified that the BERT model shows effective knowledge transfer from the self-supervised nlp tasks with a large-scale pre-training corpus. 

Based on the architecture, we propose to utilize the cell-level layout information from document images and incorporate them into the transformer encoder. First, given a set of tokens from different cells and the layout information of cells, the cell-level input embeddings are computed by summing the corresponding word embeddings, cell-level 2D-position embeddings, and original 1D-position embeddings. Then, these input embeddings are passed through a bidirectional Transformer encoder that can generate contextualized representations with an attention mechanism.

\subsection{Cell-level Input Embedding}
Given document images, we use an OCR tool to recognize text and serialize the cells (bounding boxes) from top-left to bottom-right. Each document image is represented as a sequence of cells $\{c_{1},...,c_{n}\}$, and each cell is composed of a sequence of words $c_i=\{w_{i}^{1},...,w_{i}^{m}\}$. Given the sequences of cells and words, we first introduce the method of cell-level input embedding.

\textbf{Cell-level Layout Embedding.} Unlike the position embedding that models the word position in a sequence, the 2D-position embedding aims to model the relative spatial position in a document image. To represent the spatial position of cells in scanned document images, we consider a document page as a coordinate system with the top-left origin. In this setting, the cell (bounding box) can be precisely defined by (x0, y0, x1, y1), where (x0, y0) corresponds to the top-left position, and (x1, y1) represents the bottom-right position. Therefore, we add two cell-level position embedding layers to embed x-axis features and y-axis features separately. The words $\{w_{i}^{1},...,w_{i}^{m}\}$ in i-th cell $c_i$ share the same 2D-position embeddings, which is different from the word-level 2D-position embedding in LayoutLM. As shown in Figure \ref{fig:framework}, the input tokens with the same color background are from the same cell, and the corresponding 2D-positions are also the same. In this way, StructuralLM can not only learn the layout information of cells but also know which words are from the same cell, which is better to obtain the contextual representation of cells. In addition, we keep the classic 1D-position embeddings to preserve the positional relationship of the tokens within the same cell. Finally, the cell-level layout embeddings are computed by summing the four 2D-position embeddings and the classic 1D-position embeddings.

\textbf{Input Embedding.} 
Given a sequence of cells $\{c_{1},...,c_{n}\}$, we use WordPiece \cite{Wu} to tokenize the words in the cells. The length of the text sequence is limited to ensure that the length of the final sequence is not greater than the maximum sequence length $L$. The final cell-level input embedding is the sum of the three embeddings. Word embedding represents the word itself, 1D-position embedding represents the token index, and cell-level 2D-position embedding is used to model the relative spatial position of cells in a document image.

\subsection{Pre-training StructuralLM}
We adopt two self-supervised tasks during the pre-training stage, which are described as follows.

\textbf{Masked Visual-Language Modeling.} We use the Masked Visual-Language Modeling (MVLM)\cite{layoutlm} to make the model learn the cell representation with the clues of cell-level 2D-position embeddings and text embeddings. We randomly mask some of the input tokens but keep the corresponding cell-level position embeddings, and then the model is pre-trained to predict the masked tokens. With the cell-level layout information, StructuralLM can know which words surrounding the mask token are in the same cell and which are in adjacent cells. In this way, StructuralLM not only utilizes the corresponding cell-level position information but also understands the cell-level contextual representation. Therefore, compared with the MVLM in LayoutLM, StructuralLM makes use of the cell-level layout information and predicts the mask tokens more accurately. We will compare the performance of the MVLM with the cell-level layout embeddings and word-level layout embeddings respectively in Section \ref{ablation}.

\textbf{Cell Position Classification.} In addition to the MVLM, we propose a new Cell Position Classification (CPC) task to model the relative spatial position of cells in a document. Given a set of scanned documents, this task aims to predict where the cells are in the documents. First, we split them into N areas of the same size. Then we calculate the area to which the cell belongs to through the center 2D-position of the cell. Meanwhile, some cells are randomly selected, and the 2D-positions of tokens in the selected cells are replaced with $(0;0;0;0)$. During the pre-training, a classification layer is built above the encoder outputs. This layer predicts a label $[1, N]$ of the area where the selected cell is located, and computes the cross-entropy loss. Considering the MVLM and CPC are performed simultaneously, the cells with masked tokens will not be selected for the CPC task. This prevents the model from not utilizing cell-level layout information when doing the MVLM task. We will compare the performance of different $N$ in Section \ref{config}.

\subsection{Fine-tuning}
The pre-trained StructuralLM model is fine-tuned on three document image understanding tasks, each of which contains form images. These three tasks are form understanding task, document visual question answering task, and document image classification task. For the form understanding task, StructuralLM predicts {B, I, E, S, O} tags for each token, and then uses sequential labeling to find the four types of entities including the question, answer, header, or other. For the document visual question answering task, we treat it as an extractive QA task and build a token-level classifier on the top of token representations, which is usually used in Machine Reading Comprehension (MRC)  \cite{squad,wangwei}. For the document image classification task, StructuralLM predicts the class labels using the representation of the $[CLS]$ token.

\section{Experiments}

\subsection{Pre-training Configuration}
\label{config}
\textbf{Pre-training Dataset}. Following LayoutLM, we pre-train StructuralLM on the IIT-CDIP Test Collection 1.0 \cite{Lewis}. It is a large-scale scanned document image dataset, which contains more than 6 million documents, with more than 11 million scanned document images. The pre-training dataset (IIT-CDIP Test Collection) only contains pure texts while missing their corresponding bounding boxes. Therefore, we need to re-process the scanned document images to obtain the layout information of cells. Like the pre-processing method of LayoutLM, we similarly process the dataset by using Tesseract~\footnote{https://github.com/tesseract-ocr/tesseract}, which is an open-source OCR engine. We normalize the actual coordinates to integers in the range from 0 to 1,000, and an empty bounding box $(0; 0; 0; 0)$ is attached to special tokens [CLS], [SEP] and [PAD].

\textbf{Implementation Details}. StructuralLM is based on the Transformer which consists of a 24-layer encoder with 1024 embedding/hidden size, 4096 feed-forward filter size, and 16 attention heads. To take advantage of existing pre-trained models and adapt to document image understanding tasks, we initialize the weight of StructuralLM model with the pre-trained RoBERTa\cite{roberta} large model except for the 2D-position embedding layers.

\begin{table*}[t]
\center
\begin{tabular}{ l c c c c }
\hline\hline
\textbf{Model} & \textbf{Precision} & \textbf{Recall} & \textbf{F1} & \textbf{Parameters} \\
\hline
BERT$_{BASE}$\cite{devlin-etal-2019-bert} & 0.5469 & 0.6710 & 0.6026 & 110M \\
RoBERTa$_{BASE}$ \cite{roberta} & 0.6349 & 0.6975 & 0.6648 & 125M \\
BERT$_{LARGE}$ & 0.6113 & 0.7085 & 0.6563 & 349M \\
RoBERTa$_{LARGE}$ & 0.6780 & 0.7391 & 0.7072 & 355M \\
BROS \cite{bros} & 0.8056 & 0.8188 & 0.8121 & - \\
\hline
LayoutLM$_{BASE}$\cite{layoutlm} & 0.7597 & 0.8155 & 0.7866 & 113M \\
LayoutLM$_{LARGE}$ & 0.7596 & 0.8219 & 0.7895 & 343M \\
\hline
StructuralLM$_{LARGE}$ & \textbf{0.8352} & \textbf{0.8681} & \textbf{0.8514} & 355M  \\
\hline\hline
\end{tabular}
\caption{Model accuracy (Precision, Recall, F1) on the test set of FUNSD.}
\label{table:funsd}
\vspace{-10pt}
\end{table*}

Following \citet{devlin-etal-2019-bert}, for the masked visual-language model task, we select 15\% of the input tokens for prediction. We replace these masked tokens with the $mask$ token 80\% of the time, a random token 10\% of the time, and an unchanged token 10\% of the time. Then, the model predicts the corresponding token with the cross-entropy loss. For the Bounding-box position classification task, we split the document image into N areas of the same size, and then select 15\% of the cells for prediction. We replace the 2D-positions of words in the masked cells with the $(0;0;0;0)$ 90\% of the time, and an unchanged position 10\% of the time.

StructuralLM is pre-trained on 16 NVIDIA Tesla V100 32GB GPUs for 480K steps, with each mini-batch containing 128 sequences of maximum length 512 tokens. The Adam optimizer is used with an initial learning rate of 1e-5 and a linear decay learning rate schedule.

\begin{figure}[t]
     \centering
     \includegraphics[width=0.5\textwidth]{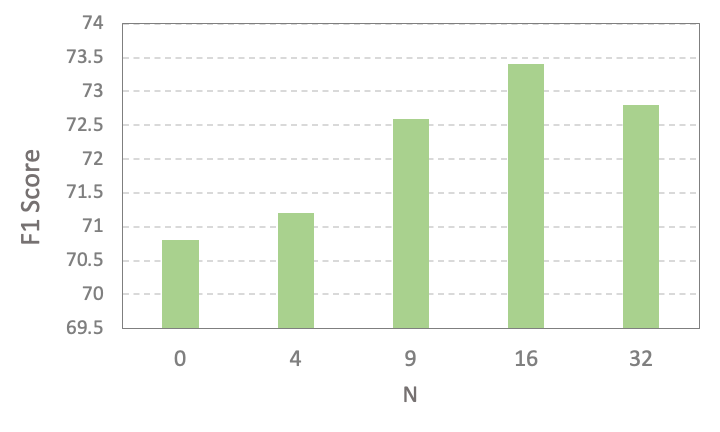}
     \caption{F1 score of StructuralLM pre-training w.r.t different hyperparameter $N$ and fine-tuning on FUNSD dataset.}
     \label{fig:N}
\end{figure}

\textbf{Hyperparameter N}. For the cell position classification task, we test the performances of StructuralLM using different hyperparameter $N$ during pre-training. Considering that the complete pre-training takes too long, we pre-train StructuralLM for 100k steps with a single GPU card to compare the performance of different $N$. As shown in Figure \ref{fig:N}, when the $N$ is set as 16, StructuralLM obtains the highest F1-score on the FUNSD dataset. Therefore, we set $N$ as 16 during the pre-training.

\subsection{Fine-tuning on Form Understanding}

We experiment with fine-tuning StructuralLM on several downstream document image understanding tasks, especially form understanding tasks. The FUNSD\cite{FUNSD} is a dataset for form understanding. It includes 199 real, fully annotated, scanned forms with 9,707 semantic entities and 31,485 words. The 199 scanned forms are split into 149 for training and 50 for testing. The FUNSD dataset is suitable for a variety of tasks, where we just fine-tuning StructuralLM on semantic entity labeling. Specifically, each word in the dataset is assigned to a semantic entity label from a set of four predefined categories: question, answer, header, or other. Following the previous works, we also use the word-level F1 score as the evaluation metric. 

We fine-tune the pre-trained StructuralLM on the FUNSD training set for 25 epochs. We set the batch size to 4, the learning rate to 1e-5. The other hyperparameters are kept the same as pre-training. 

Table \ref{table:funsd} presents the experimental results on the FUNSD test set. StructuralLM achieves better performance than all pre-training models. First, we compare the StructuralLM model with two SOTA text-only pre-trained models: BERT and RoBERTa\cite{roberta}. RoBERTa outperforms the BERT model by a large margin in terms of the BASE and LARGE settings. Compared with the text-only models, the text+layout model LayoutLM brings significant performance improvement. The best performance is achieved by StructuralLM, where an improvement of 6\% F1 point compared with LayoutLM under the same model size.
By consistently outperforming the pre-training methods, StructuralLM confirms its effectiveness in leveraging cell-level layout information for form understanding.

\subsection{Fine-tuning on Document Visual QA}

\begin{table}[t]
\center
\begin{tabular}{ l c c }
\hline\hline
\multirow{2}{*}{\textbf{Model}} & \textbf{ANLS} & \textbf{ANLS}\\
& \textbf{Test set} & \textbf{Form\&Table}\\
\hline
BERT$_{BASE}$ & 0.6372 & - \\
RoBERTa$_{BASE}$ & 0.6642  & - \\
BERT$_{LARGE}$ & 0.6745 & - \\
RoBERTa$_{LARGE}$ & 0.6952 & - \\
\hline
LayoutLM$_{BASE}$ & 0.6979 & 0.7012 \\
LayoutLM$_{LARGE}$ & 0.7259 & 0.7203 \\
\hline
StructuralLM$_{LARGE}$ & \textbf{0.8394} & \textbf{0.8610}  \\
\hline\hline
\end{tabular}
\caption{Average Normalized Levenshtein Similarity (ANLS) score on the DocVQA test set and the Form\&Table subset from the test set.}
\label{table:docvqa}
\vspace{-10pt}
\end{table}

DocVQA \cite{DocVQA} is a VQA dataset on the scanned document understanding field. The objective of this task is to answer questions asked on a document image. The images provided are sourced from the documents hosted at the Industry Documents Library, maintained by the UCSF. It consists of 12,000 pages from a variety of documents including forms, tables, etc. These pages are manually labeled with 50,000 question-answer pairs, which are split into the training set, validation set and test set with a ratio of about 8:1:1. The dataset is organized as a set of triples (page image, questions, answers). The official provides the OCR results of the page images, and there is no objection to using other OCR recognition tools. Our experiment is based on the official OCR results. The task is evaluated using an edit distance based metric ANLS (aka average normalized Levenshtein similarity). Results on the test set are provided by the official evaluation site.

We fine-tune the pre-trained StructuralLM on the DocVQA train set and validation set for 5 epochs. We set the batch size to 8, the learning rate to 1e-5.

Table \ref{table:docvqa} shows the Average Normalized Levenshtein Similarity (ANLS) scores on the DocVQA test set. We still compare the StructuralLM model with the text-only models and the text-layout model. Compared with LayoutLM, StructuralLM achieved an improvement of over 11\% ANLS point under the same model size. In addition, we also compare the Form\&Table subset from the test set. StructuralLM achieved an improvement of over 14\% ANLS point, which shows that StructuralLM can learn better on form and table understanding.

\begin{table}[t]
\center
\begin{tabular}{ l c c }
\hline\hline
\textbf{Model} & \textbf{Acc} & \textbf{Params} \\
\hline
BERT$_{BASE}$ & 89.81\% & 110M \\
RoBERTa$_{BASE}$ & 90.06\% & 125M \\
BERT$_{LARGE}$ & 89.92\% & 349M \\
RoBERTa$_{LARGE}$ & 90.11\% & 355M \\
\hline
VGG-16$^a$ & 90.97\% & - \\
Stacked CNN Single$^b$ & 91.11\% & - \\
Stacked CNN Ensemble$^b$ & 92.21\% & - \\
InceptionResNetV2$^c$ & 92.63\% & - \\
LadderNet$^d$ & 92.77\% & -  \\
Multimodal Single$^e$ & 93.03\% & - \\
Multimodal Ensemble$^e$ & 93.07\% & - \\
\hline
LayoutLM$_{BASE}$ & 94.42\% & 113M \\
LayoutLM$_{LARGE}$ & 94.43\% & 390M \\
\hline
StructuralLM$_{LARGE}$ & \textbf{96.08\%} & 355M  \\
\hline\hline
\end{tabular}
\caption{Classification accuracy on the RVL-CDIP test set. $^a$~\cite{VGG};$^b$~\cite{Stacked};$^c$~\cite{inception};$d$~\cite{LadderNet};$^e$~\cite{Multimodal}}
\label{table:rvlcdip}
\vspace{-10pt}
\end{table}

\subsection{Fine-tuning on Document Classification}

Finally, we evaluate the document image classification task using the RVL-CDIP dataset\cite{harley2015evaluation}. It consists of 400,000 grayscale images in 16 classes, with 25,000 images per class. There are 320,000 images for the training set, 40,000 images for the validation set, and 40,000 images for the test set. A multi-class single-label classification task is defined on RVL-CDIP, including letter, form, invoice, etc. The evaluation metric is the overall classification accuracy. Text and layout information is extracted by Tesseract OCR.

We fine-tune the pre-trained StructuralLM on the RVL-CDIP train set for 20 epochs. We set the batch size to 8, the learning rate to 1e-5.

Different from other natural images, the document images are texts in a variety of layouts. As shown in Table \ref{table:rvlcdip}, image-based classification models\cite{VGG,Stacked,inception} with pre-training perform much better than the text-based models, which illustrates that text information is not sufficient for this task and it still needs layout information. The experiment results show that the text-layout model LayoutLM outperforms the image-based approaches and text-based models. Incorporating the cell-level layout information, StructuralLM achieves a new state-of-the-art result with an improvement of over 1.5\% accuracy point.

\subsection{Ablation Study}

\begin{figure*}[t]
     \centering
     \includegraphics[width=0.95\textwidth]{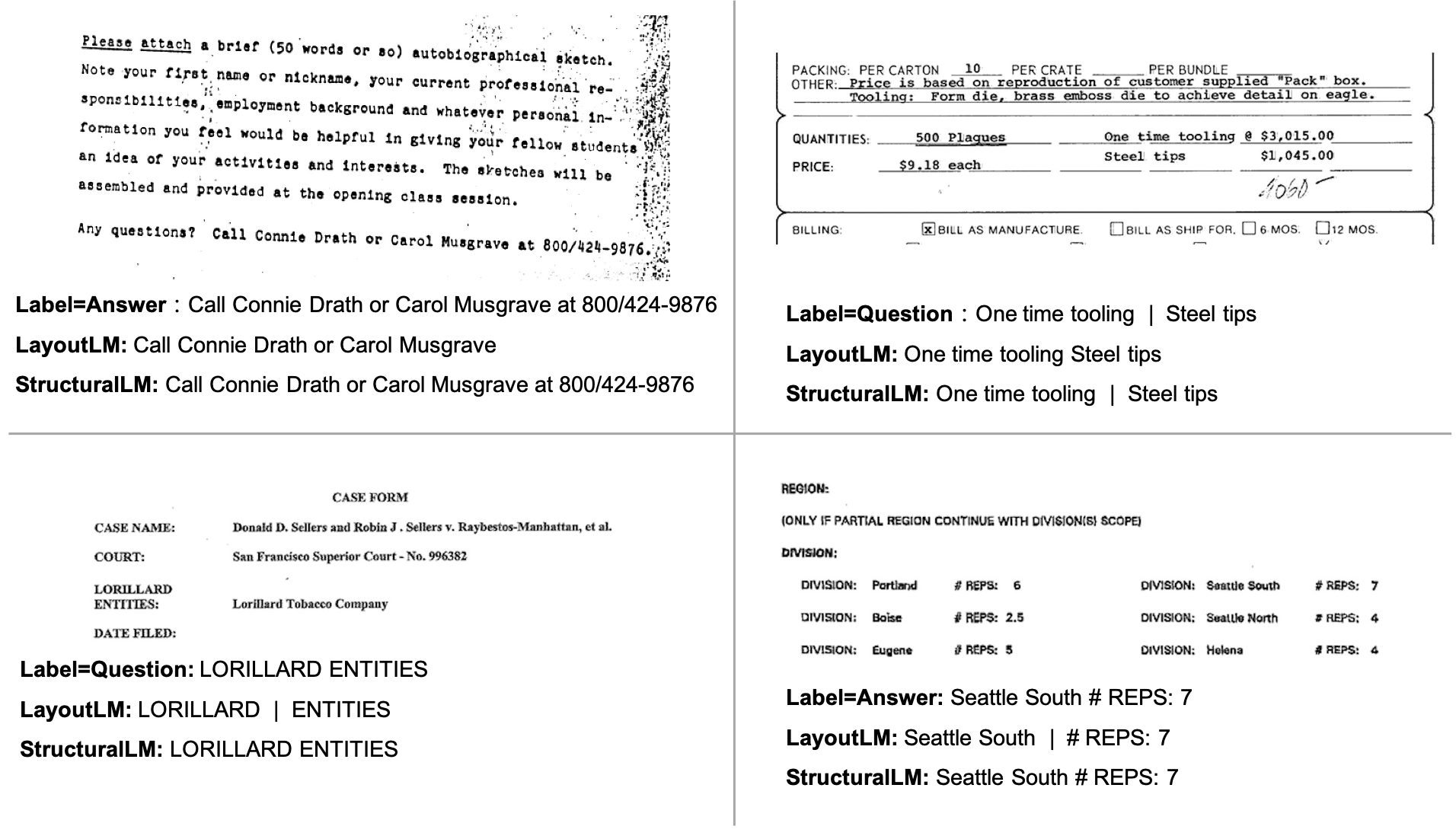}
     \caption{Examples of the output of LayoutLM and StructuralLM on the FUNSD dataset. The division of $|$ means that the two phrases are independent labels.}
     \label{fig:case_study}
\end{figure*}

\label{ablation}

We conduct ablation studies to assess the individual contribution of every component in StructuralLM. Table \ref{table:ablation} reports the results of full StructuralLM and its ablations on the test set of FUNSD form understanding task. First, we evaluate how much the cell-level layout embedding contributes to form understanding by removing it from StructuralLM pre-training. 
\begin{table}[t]
\center
\begin{tabular}{ l | c}
\hline\hline
{\textbf{Ablation}} & \textbf{F1}\\
\hline\hline
\textbf{StructuralLM} & \textbf{0.8514} \\
\hline
w/o cell-level layout embedding & 0.8024  \\
\hline
w/o cell position classification & 0.8125  \\
\hline
w/o pre-training & 0.7072 \\
\hline
LayoutLM \\
w/ word-level layout embedding & 0.7895    \\
\hline\hline
\end{tabular}
\caption{Ablation tests of StructuralLM on the FUNSD form understanding task.}
\label{table:ablation}
\vspace{-8pt}
\end{table}
This ablation results in a drop from 0.8514 to 0.8024 on F1 score, demonstrating the important role of the cell-level layout embedding. To study the effect of the cell position classification task in StructuralLM, we ablate it and the F1 score significantly drops from 0.8514 to 0.8125. Finally, we study the significance of full StructuralLM pre-training. Over 15\% of performance degradation resulted from ablating pre-training clearly demonstrates the power of StructuralLM in leveraging an unlabeled corpus for downstream form understanding tasks.

\begin{figure}[t]
     \centering
     \includegraphics[width=0.5\textwidth]{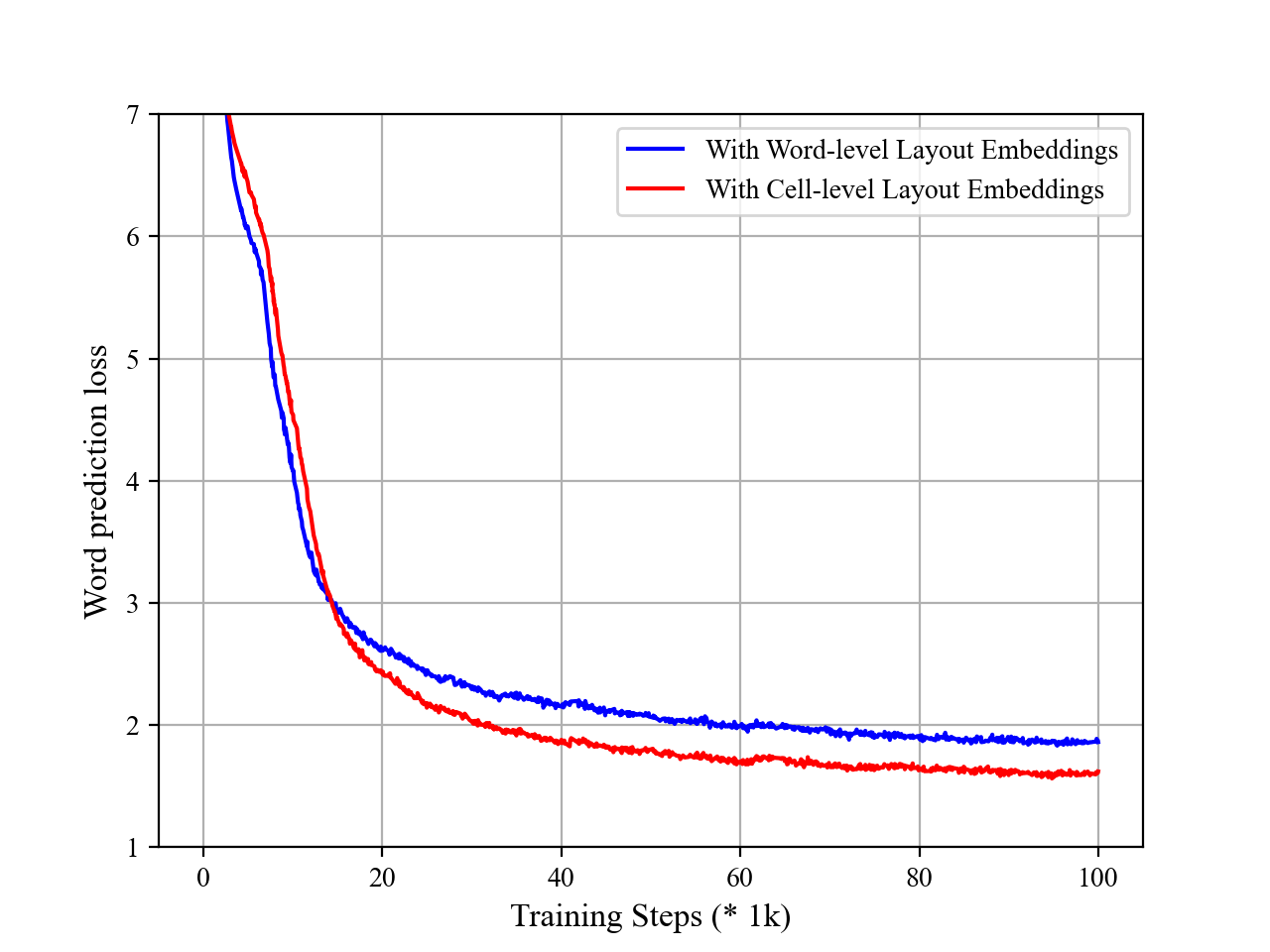}
     \caption{Loss of word prediction over the number of pre-training steps based on different layout embeddings.}
     \label{fig:compare}
\end{figure}

Actually, after ablating the cell position classification, the biggest difference between StructuralLM and LayoutLM is cell-level 2D-position embeddings or word-level 2D-position embeddings. The results show that StructuralLM with cell-level 2D-position embeddings performs better than LayoutLM with word-level position embeddings with an improvement of over 2\% F1-score point (from 0.7895 to 0.8125). Furthermore, we compare the performance of the MVLM with cell-level layout embeddings and word-level layout embeddings respectively. As shown in Figure \ref{fig:compare}, the results show that under the same pre-training settings, the MVLM training loss with cell-level 2D-position embeddings can converge lower.

\subsection{Case Study}
The motivation behind StructuralLM is to jointly exploit cell and layout information across scanned document images. As stated above, compared with LayoutLM, StructuralLM improves interactions between cells and layout information. To verify this, we show some examples of the output of LayoutLM and StructuralLM on the FUNSD test set, as shown in Figure \ref{fig:case_study}. Take the image on the top-left of Figure \ref{fig:case_study} as an example. In this example, the model needs to label "Call Connie Drath or Carol Musgrave at 800/424-9876" with the \emph{Answer} entity. The result of LayoutLM missed "at 800/424-9876". Actually, all the tokens of this \emph{Answer} entity are from the same cell. Therefore, StructuralLM predicts the correct result with the understanding of cell-level layout information. These examples show that StructuralLM predicts the entities more accurately with the cell-level layout information. The same results can be observed in the Figure \ref{fig:case_study}.

\section{Related Work}
\subsection{Machine Learning Approaches}
Statistical machine learning approaches \cite{Marinai,shilman} became the mainstream for document segmentation tasks during the past decade. \cite{shilman} consider the layout information of a document as a parsing problem. They use a grammar-based loss function to globally search the optimal parsing tree, and utilize a machine learning approach to select features and train all parameters during the parsing process. In addition, most efforts have been devoted to the recognition of isolated handwritten and printed characters with widely recognized successful results. For machine learning approaches\cite{shilman,Wei2013EvaluationOS}, they are usually time-consuming to design manually features and difficult to obtain a high-level abstract semantic context. In addition, these methods usually relied on visual cues but ignored textual information.
\subsection{Deep Learning Approaches}
Nowadays, deep learning methods have become the mainstream for many machine learning problems\cite{yang2017,Oliveira,Katti,soto}. \cite{yang2017} propose a pixel-by-pixel classification to solve the document semantic structure extraction problem. Specifically, they propose a multi-modal neural network that considers visual and textual information, while this work is an end-to-end approach. \cite{Katti} first propose a fully convolutional encoder-decoder network to predict a segmentation mask and bounding boxes. In this way, the model significantly outperforms approaches based on sequential text or document images. In addition,  \cite{soto} incorporate contextual information into the Faster R-CNN model. They involve the inherently localized nature of article contents to improve region detection performance.
\subsection{Pre-training Approaches}
In recent years, self-supervised pre-training has achieved great success in natural language understanding (NLU) and a wide range of NLP tasks\cite{devlin-etal-2019-bert,liu2019roberta,wang2019structbert}. \cite{devlin-etal-2019-bert} introduced BERT, a new language representation model, which is designed to pre-train deep bidirectional representations based on the large-scale unsupervised corpus. It can be fine-tuned with just one additional output layer to create state-of-the-art models for a wide range of NLP tasks. Inspired by the development of the pre-trained language models in various NLP tasks, recent studies on document image pre-training \cite{zhang2020trie,layoutlm} do have pushed the limits of a variety of document image understanding tasks, which learn the interaction between text and layout information across scanned document images. \cite{layoutlm} propose LayoutLM, which is a simple but effective pre-training method of text and layout for the document image understanding tasks. By incorporating the visual information into the fine-tuning stage, LayoutLM achieves new state-of-the-art results in several downstream tasks.

\section{Conclusion}
In this paper, we propose StructuralLM, a novel structural pre-training approach on large unlabeled documents. It is built upon an extension of the Transformer encoder, and jointly exploit cell and layout information from scanned documents. 

Different from previous pre-trained models, StructuralLM uses cell-level 2D-position embeddings with tokens in the cell sharing the same 2D-position. This makes StructuralLM aware of which words are from the same cell, and thus enables the model to derive representation for the cells. We propose a new pre-training objective called cell position classification. In this way, StructuralLM is capable of learning the interactions between cells and layout. We conduct experiments on three benchmark datasets publicly available, and StructuralLM outperforms strong baselines and achieves new state-of-the-art results in the downstream tasks.

\bibliographystyle{acl_natbib}
\bibliography{acl2021}

\begin{thebibliography}{26}
\expandafter\ifx\csname natexlab\endcsname\relax\def\natexlab#1{#1}\fi

\bibitem[{Afzal et~al.(2017)Afzal, K{\"o}lsch, Ahmed, and Liwicki}]{VGG}
Muhammad~Zeshan Afzal, Andreas K{\"o}lsch, Sheraz Ahmed, and Marcus Liwicki.
  2017.
\newblock Cutting the error by half: Investigation of very deep cnn and
  advanced training strategies for document image classification.
\newblock In \emph{2017 14th IAPR International Conference on Document Analysis
  and Recognition (ICDAR)}, volume~1, pages 883--888. IEEE.

\bibitem[{{Borges Oliveira} and {Viana}(2017)}]{Oliveira}
D.~A. {Borges Oliveira} and M.~P. {Viana}. 2017.
\newblock \href {https://doi.org/10.1109/ICCVW.2017.142} {Fast cnn-based
  document layout analysis}.
\newblock In \emph{2017 IEEE International Conference on Computer Vision
  Workshops (ICCVW)}, pages 1173--1180.

\bibitem[{Das et~al.(2018)Das, Roy, Bhattacharya, and Parui}]{Stacked}
Arindam Das, Saikat Roy, Ujjwal Bhattacharya, and Swapan~K Parui. 2018.
\newblock Document image classification with intra-domain transfer learning and
  stacked generalization of deep convolutional neural networks.
\newblock In \emph{2018 24th International Conference on Pattern Recognition
  (ICPR)}, pages 3180--3185. IEEE.

\bibitem[{Dauphinee et~al.(2019)Dauphinee, Patel, and Rashidi}]{Multimodal}
Tyler Dauphinee, Nikunj Patel, and Mohammad Rashidi. 2019.
\newblock Modular multimodal architecture for document classification.
\newblock \emph{arXiv preprint arXiv:1912.04376}.

\bibitem[{Devlin et~al.(2019)Devlin, Chang, Lee, and
  Toutanova}]{devlin-etal-2019-bert}
Jacob Devlin, Ming-Wei Chang, Kenton Lee, and Kristina Toutanova. 2019.
\newblock \href {https://doi.org/10.18653/v1/N19-1423} {{BERT}: Pre-training of
  deep bidirectional transformers for language understanding}.
\newblock In \emph{Proceedings of the 2019 Conference of the North {A}merican
  Chapter of the Association for Computational Linguistics: Human Language
  Technologies, Volume 1 (Long and Short Papers)}, pages 4171--4186,
  Minneapolis, Minnesota. Association for Computational Linguistics.

\bibitem[{Harley et~al.(2015)Harley, Ufkes, and
  Derpanis}]{harley2015evaluation}
Adam~W Harley, Alex Ufkes, and Konstantinos~G Derpanis. 2015.
\newblock Evaluation of deep convolutional nets for document image
  classification and retrieval.
\newblock In \emph{2015 13th International Conference on Document Analysis and
  Recognition (ICDAR)}, pages 991--995. IEEE.

\bibitem[{Hong et~al.(2021)Hong, Kim, Ji, Hwang, Nam, and Park}]{bros}
Teakgyu Hong, DongHyun Kim, Mingi Ji, Wonseok Hwang, Daehyun Nam, and Sungrae
  Park. 2021.
\newblock \href {https://openreview.net/forum?id=punMXQEsPr0} {{\{}BROS{\}}: A
  pre-trained language model for understanding texts in document}.

\bibitem[{Jaume et~al.(2019)Jaume, Ekenel, and Thiran}]{FUNSD}
Guillaume Jaume, Hazim~Kemal Ekenel, and Jean{-}Philippe Thiran. 2019.
\newblock \href {http://arxiv.org/abs/1905.13538} {{FUNSD:} {A} dataset for
  form understanding in noisy scanned documents}.
\newblock \emph{CoRR}, abs/1905.13538.

\bibitem[{Katti et~al.(2018)Katti, Reisswig, Guder, Brarda, Bickel, H{\"o}hne,
  and Faddoul}]{Katti}
Anoop~Raveendra Katti, Christian Reisswig, Cordula Guder, Sebastian Brarda,
  Steffen Bickel, Johannes H{\"o}hne, and Jean~Baptiste Faddoul. 2018.
\newblock Chargrid: Towards understanding 2d documents.
\newblock \emph{arXiv preprint arXiv:1809.08799}.

\bibitem[{Lewis et~al.(2006)Lewis, Agam, Argamon, Frieder, Grossman, and
  Heard}]{Lewis}
D.~Lewis, G.~Agam, S.~Argamon, O.~Frieder, D.~Grossman, and J.~Heard. 2006.
\newblock \href {https://doi.org/10.1145/1148170.1148307} {Building a test
  collection for complex document information processing}.
\newblock SIGIR '06, page 665–666, New York, NY, USA. Association for
  Computing Machinery.

\bibitem[{Liu et~al.(2019{\natexlab{a}})Liu, Ott, Goyal, Du, Joshi, Chen, Levy,
  Lewis, Zettlemoyer, and Stoyanov}]{liu2019roberta}
Yinhan Liu, Myle Ott, Naman Goyal, Jingfei Du, Mandar Joshi, Danqi Chen, Omer
  Levy, Mike Lewis, Luke Zettlemoyer, and Veselin Stoyanov. 2019{\natexlab{a}}.
\newblock Roberta: A robustly optimized bert pretraining approach.
\newblock \emph{arXiv preprint arXiv:1907.11692}.

\bibitem[{Liu et~al.(2019{\natexlab{b}})Liu, Ott, Goyal, Du, Joshi, Chen, Levy,
  Lewis, Zettlemoyer, and Stoyanov}]{roberta}
Yinhan Liu, Myle Ott, Naman Goyal, Jingfei Du, Mandar Joshi, Danqi Chen, Omer
  Levy, Mike Lewis, Luke Zettlemoyer, and Veselin Stoyanov. 2019{\natexlab{b}}.
\newblock \href {http://arxiv.org/abs/1907.11692} {Roberta: A robustly
  optimized bert pretraining approach}.
\newblock Cite arxiv:1907.11692.

\bibitem[{{Marinai} et~al.(2005){Marinai}, {Gori}, and {Soda}}]{Marinai}
S.~{Marinai}, M.~{Gori}, and G.~{Soda}. 2005.
\newblock \href {https://doi.org/10.1109/TPAMI.2005.4} {Artificial neural
  networks for document analysis and recognition}.
\newblock \emph{IEEE Transactions on Pattern Analysis and Machine
  Intelligence}, 27(1):23--35.

\bibitem[{Mathew et~al.(2020)Mathew, Karatzas, Manmatha, and Jawahar}]{DocVQA}
M.~Mathew, Dimosthenis Karatzas, R.~Manmatha, and C.~Jawahar. 2020.
\newblock Docvqa: A dataset for vqa on document images.
\newblock \emph{ArXiv}, abs/2007.00398.

\bibitem[{Rajpurkar et~al.(2016)Rajpurkar, Zhang, Lopyrev, and Liang}]{squad}
Pranav Rajpurkar, Jian Zhang, Konstantin Lopyrev, and Percy Liang. 2016.
\newblock \href {https://doi.org/10.18653/v1/D16-1264} {{SQ}u{AD}: 100,000+
  questions for machine comprehension of text}.
\newblock In \emph{Proceedings of the 2016 Conference on Empirical Methods in
  Natural Language Processing}, pages 2383--2392, Austin, Texas. Association
  for Computational Linguistics.

\bibitem[{Sarkhel and Nandi(2019)}]{LadderNet}
Ritesh Sarkhel and Arnab Nandi. 2019.
\newblock \href {https://doi.org/10.24963/ijcai.2019/466} {Deterministic
  routing between layout abstractions for multi-scale classification of
  visually rich documents}.
\newblock In \emph{Proceedings of the Twenty-Eighth International Joint
  Conference on Artificial Intelligence, {IJCAI-19}}, pages 3360--3366.
  International Joint Conferences on Artificial Intelligence Organization.

\bibitem[{{Shilman} et~al.(2005){Shilman}, {Liang}, and {Viola}}]{shilman}
M.~{Shilman}, P.~{Liang}, and P.~{Viola}. 2005.
\newblock \href {https://doi.org/10.1109/ICCV.2005.140} {Learning nongenerative
  grammatical models for document analysis}.
\newblock In \emph{Tenth IEEE International Conference on Computer Vision
  (ICCV'05) Volume 1}, volume~2, pages 962--969 Vol. 2.

\bibitem[{Soto and Yoo(2019)}]{soto}
Carlos Soto and Shinjae Yoo. 2019.
\newblock \href {https://doi.org/10.18653/v1/D19-1348} {Visual detection with
  context for document layout analysis}.
\newblock In \emph{Proceedings of the 2019 Conference on Empirical Methods in
  Natural Language Processing and the 9th International Joint Conference on
  Natural Language Processing (EMNLP-IJCNLP)}, pages 3464--3470, Hong Kong,
  China. Association for Computational Linguistics.

\bibitem[{Szegedy et~al.(2017)Szegedy, Ioffe, Vanhoucke, and Alemi}]{inception}
Christian Szegedy, Sergey Ioffe, Vincent Vanhoucke, and Alexander Alemi. 2017.
\newblock Inception-v4, inception-resnet and the impact of residual connections
  on learning.
\newblock In \emph{Proceedings of the AAAI Conference on Artificial
  Intelligence}, volume~31.

\bibitem[{Wang et~al.(2019)Wang, Bi, Yan, Wu, Bao, Xia, Peng, and
  Si}]{wang2019structbert}
Wei Wang, Bin Bi, Ming Yan, Chen Wu, Zuyi Bao, Jiangnan Xia, Liwei Peng, and
  Luo Si. 2019.
\newblock Structbert: Incorporating language structures into pre-training for
  deep language understanding.
\newblock \emph{arXiv preprint arXiv:1908.04577}.

\bibitem[{Wang et~al.(2018)Wang, Yan, and Wu}]{wangwei}
Wei Wang, Ming Yan, and Chen Wu. 2018.
\newblock \href {https://doi.org/10.18653/v1/P18-1158} {Multi-granularity
  hierarchical attention fusion networks for reading comprehension and question
  answering}.
\newblock In \emph{Proceedings of the 56th Annual Meeting of the Association
  for Computational Linguistics (Volume 1: Long Papers)}, pages 1705--1714,
  Melbourne, Australia. Association for Computational Linguistics.

\bibitem[{Wei et~al.(2013)Wei, Baechler, Slimane, and
  Ingold}]{Wei2013EvaluationOS}
H.~Wei, M.~Baechler, F.~Slimane, and R.~Ingold. 2013.
\newblock Evaluation of svm, mlp and gmm classifiers for layout analysis of
  historical documents.
\newblock \emph{2013 12th International Conference on Document Analysis and
  Recognition}, pages 1220--1224.

\bibitem[{Wu et~al.(2016)Wu, Schuster, Chen, Le, Norouzi, Macherey, Krikun,
  Cao, Gao, Macherey, Klingner, Shah, Johnson, Liu, Łukasz Kaiser, Gouws,
  Kato, Kudo, Kazawa, Stevens, Kurian, Patil, Wang, Young, Smith, Riesa,
  Rudnick, Vinyals, Corrado, Hughes, and Dean}]{Wu}
Yonghui Wu, Mike Schuster, Zhifeng Chen, Quoc~V. Le, Mohammad Norouzi, Wolfgang
  Macherey, Maxim Krikun, Yuan Cao, Qin Gao, Klaus Macherey, Jeff Klingner,
  Apurva Shah, Melvin Johnson, Xiaobing Liu, Łukasz Kaiser, Stephan Gouws,
  Yoshikiyo Kato, Taku Kudo, Hideto Kazawa, Keith Stevens, George Kurian,
  Nishant Patil, Wei Wang, Cliff Young, Jason Smith, Jason Riesa, Alex Rudnick,
  Oriol Vinyals, Greg Corrado, Macduff Hughes, and Jeffrey Dean. 2016.
\newblock \href {http://arxiv.org/abs/1609.08144} {Google's neural machine
  translation system: Bridging the gap between human and machine translation}.
\newblock \emph{CoRR}, abs/1609.08144.

\bibitem[{Xu et~al.(2019)Xu, Li, Cui, Huang, Wei, and Zhou}]{layoutlm}
Yiheng Xu, Minghao Li, Lei Cui, Shaohan Huang, Furu Wei, and Ming Zhou. 2019.
\newblock \href {http://arxiv.org/abs/1912.13318} {Layoutlm: Pre-training of
  text and layout for document image understanding}.
\newblock \emph{CoRR}, abs/1912.13318.

\bibitem[{Yang et~al.(2017)Yang, Yumer, Asente, Kraley, Kifer, and
  Lee~Giles}]{yang2017}
Xiao Yang, Ersin Yumer, Paul Asente, Mike Kraley, Daniel Kifer, and
  C~Lee~Giles. 2017.
\newblock Learning to extract semantic structure from documents using
  multimodal fully convolutional neural networks.
\newblock In \emph{Proceedings of the IEEE Conference on Computer Vision and
  Pattern Recognition}, pages 5315--5324.

\bibitem[{Zhang et~al.(2020)Zhang, Xu, Cheng, Pu, Lu, Qiao, Niu, and
  Wu}]{zhang2020trie}
Peng Zhang, Yunlu Xu, Zhanzhan Cheng, Shiliang Pu, Jing Lu, Liang Qiao, Yi~Niu,
  and Fei Wu. 2020.
\newblock Trie: End-to-end text reading and information extraction for document
  understanding.
\newblock In \emph{Proceedings of the 28th ACM International Conference on
  Multimedia}, pages 1413--1422.

\end{thebibliography}

%\appendix
\end{document}